\documentclass{ieeeaccess}
\usepackage{cite}
\usepackage{multirow}%
\usepackage{amsmath,amssymb,amsfonts}
\usepackage{xcolor}%
\usepackage{graphicx}
\usepackage{textcomp}
\usepackage{amsmath}
\usepackage{booktabs}
\usepackage{algorithmicx}%
\usepackage{algpseudocode}%
\usepackage{listings}%
\usepackage{float}% for [H] placement
\usepackage{url}
\usepackage{kotex}

\newcommand{\comb}[2]{{}_{#1}\mathrm{C}_{#2}}

\def\BibTeX{{\rm B\kern-.05em{\sc i\kern-.025em b}\kern-.08em
    T\kern-.1667em\lower.7ex\hbox{E}\kern-.125emX}}
\begin{document}
\history{Date of publication xxxx 00, 0000, date of current version xxxx 00, 0000.}
\doi{10.1109/ACCESS.2023.0322000}

\title{Synergy-CLIP: Extending CLIP with Multi-modal Integration for Robust Representation Learning}
\author{
\uppercase{Sangyeon Cho}\authorrefmark{1},
\uppercase{Jangyeong Jeon}\authorrefmark{1},
\uppercase{Mingi Kim}\authorrefmark{1},
and \uppercase{Junyeong Kim}\authorrefmark{1}
}
\address[1]{Department of Artificial Intelligence, Chung-Ang University, Seoul 06974, Republic of Korea}
\tfootnote{This work was partly supported by the National Research Foundation of Korea(NRF) grant funded by the Korea government(MSIT) (RS-2023-00253782) and partly supported by Institute of Information \& communications Technology Planning \& Evaluation (IITP) grant funded by the Korea government(MSIT) (No. 2021-0-01341, Artificial Intelligence Graduate School Program, Chung-Ang University)}

\markboth
{S. Cho \headeretal: Synergy-CLIP: Extending CLIP with Multi-modal Integration for Robust Representation Learning}
{S. Cho \headeretal: Synergy-CLIP: Extending CLIP with Multi-modal Integration for Robust Representation Learning}

\corresp{Corresponding author: Junyeong Kim (e-mail: junyeongkim@cau.ac.kr).}

\begin{abstract}
Multi-modal representation learning has become a pivotal area in artificial intelligence, enabling the integration of diverse modalities such as vision, text, and audio to solve complex problems. However, existing approaches predominantly focus on bimodal interactions, such as image-text pairs, which limits their ability to fully exploit the richness of multi-modal data. Furthermore, the integration of modalities in equal-scale environments remains underexplored due to the challenges of constructing large-scale, balanced datasets. In this study, we propose Synergy-CLIP, a novel framework that extends the contrastive language-image pre-training (CLIP) architecture to enhance multi-modal representation learning by integrating visual, textual, and audio modalities. Unlike existing methods that focus on adapting individual modalities to vanilla-CLIP, Synergy-CLIP aligns and captures latent information across three modalities equally. To address the high cost of constructing large-scale multi-modal datasets, we introduce VGG-sound+, a triple-modal dataset designed to provide equal-scale representation of visual, textual, and audio data. Synergy-CLIP is validated on various downstream tasks, including zero-shot classification, where it outperforms existing baselines. Additionally, we introduce a missing modality reconstruction task, demonstrating Synergy-CLIP’s ability to extract synergy among modalities in realistic application scenarios. These contributions provide a robust foundation for advancing multi-modal representation learning and exploring new research directions.
The code is available at \url{https://github.com/JoSangYeon/Synergy-CLIP}
\end{abstract}


\titlepgskip=-21pt

\maketitle

%%%%%%%%%%%%%%%%%%%%%%%%%%%%%%%%%%%%%%%%%%%%%%%%%%%%%%%%%%%%%%%%%%%%%%%%

\section{Introduction} \label{sec:introduction}
\PARstart{R}{ecent} progress in the domain of artificial intelligence (AI) has revealed multi-modal learning as an innovative paradigm \cite{nagrani2021attention,zhou2021vmloc,han2022trusted} that utilizes information spread over various interfaces, such as visual and auditory modalities. Multi-modal AI leverages diverse input data (i.e., multi-modal data) to solve complex problems, thereby overcoming the inherent limitations of models trained on single-modality data and exhibiting superior performance \cite{taleb2021multimodal,rahate2022multimodal}. 
Recent ongoing studies \cite{qu2021efficient} have focused on the fusion and utilization of various modalities, including discussions on the effectiveness and efficiency of modeling methodologies, to facilitate learning and improve them in deep learning applications such as computer vision \cite{bayoudh2022survey,chai2022deep} and natural language processing \cite{xu2023multimodal,rahate2022multimodal}. In addition, a groundbreaking approach known as contrastive language-image pre-training (CLIP) \cite{radford2021learning} has been introduced. CLIP utilizes large datasets for training by combining images and text, and efficiently links vision and language models. It aligns the representations in the two modalities while matching images to relevant text descriptions, focusing on understanding the interactions between the input modalities. The innovative approach and robust performance of CLIP have contributed significantly to the advancement of multi-modal representation learning, marking an important milestone in subsequent research \cite{elizalde2023clap,zhang2022pointclip,guzhov2022audioclip}.

% However, research on the CLIP model has primarily focused on bimodal interactions \cite{zhang2022pointclip,zhu2021deep,liang2022mind,jiang2023understanding}, such as image-text \cite{radford2021learning} and audio-text data \cite{elizalde2023clap}. Attempts have been made to adapt \cite{guzhov2022audioclip, ruan2023accommodating, wu2022wav2clip} new modalities to existing bimodal frameworks; however, expanding this research to learn and represent all modalities equally remains challenging. These limitations stem from the high cost \cite{zhan2021comparative, liu2020generative} associated with the annotation and acquisition of large-scale tri-modal datasets for pre-training. Therefore, a notable lack of research utilizes three or more modalities in cross-domain contexts \cite{liang2021multibench,guzhov2022audioclip,xue2023ulip}. However, to endow models with the capability to accommodate and integrate diverse information like humans, these impediments must be overcome and the scope of research should be expanded to encompass multiple modalities. Therefore, in this study, we emphasize the importance of extending the research to include tri-modal representation learning, to develop models capable of processing and understanding complex information in a human-like manner. We believe this would endow models with the ability to accommodate and integrate diverse information in a human-like manner.
However, follow-up research on the CLIP model remains largely limited to exploring bimodal interactions \cite{zhang2022pointclip,zhu2021deep,liang2022mind,jiang2023understanding}, such as image-text \cite{radford2021learning} and audio-text data \cite{elizalde2023clap}. The addition of new modalities primarily takes the form of adapting individual modalities within a bimodal framework \cite{guzhov2022audioclip, ruan2023accommodating, wu2022wav2clip}, making it challenging to pursue research that learns and represents all modalities on an equal footing. These limitations stem from the high cost \cite{zhan2021comparative, liu2020generative} of annotating and acquiring large-scale tri-modal datasets for pre-training. However, to endow models with the ability to process and integrate diverse information like humans, there is a pressing need for research that encompasses multiple modalities with equal representation. Therefore, we underscore the importance of moving beyond bimodal approaches to emphasize the need for datasets that support tri-modal representation learning, facilitating the development of models that can process and understand complex information in a human-like manner. We envision this expansion will enable models to absorb and integrate diverse types of information, thus approaching human-like cognitive capabilities.

\begin{figure}[!t]
\centering
\includegraphics[width=0.975\columnwidth]{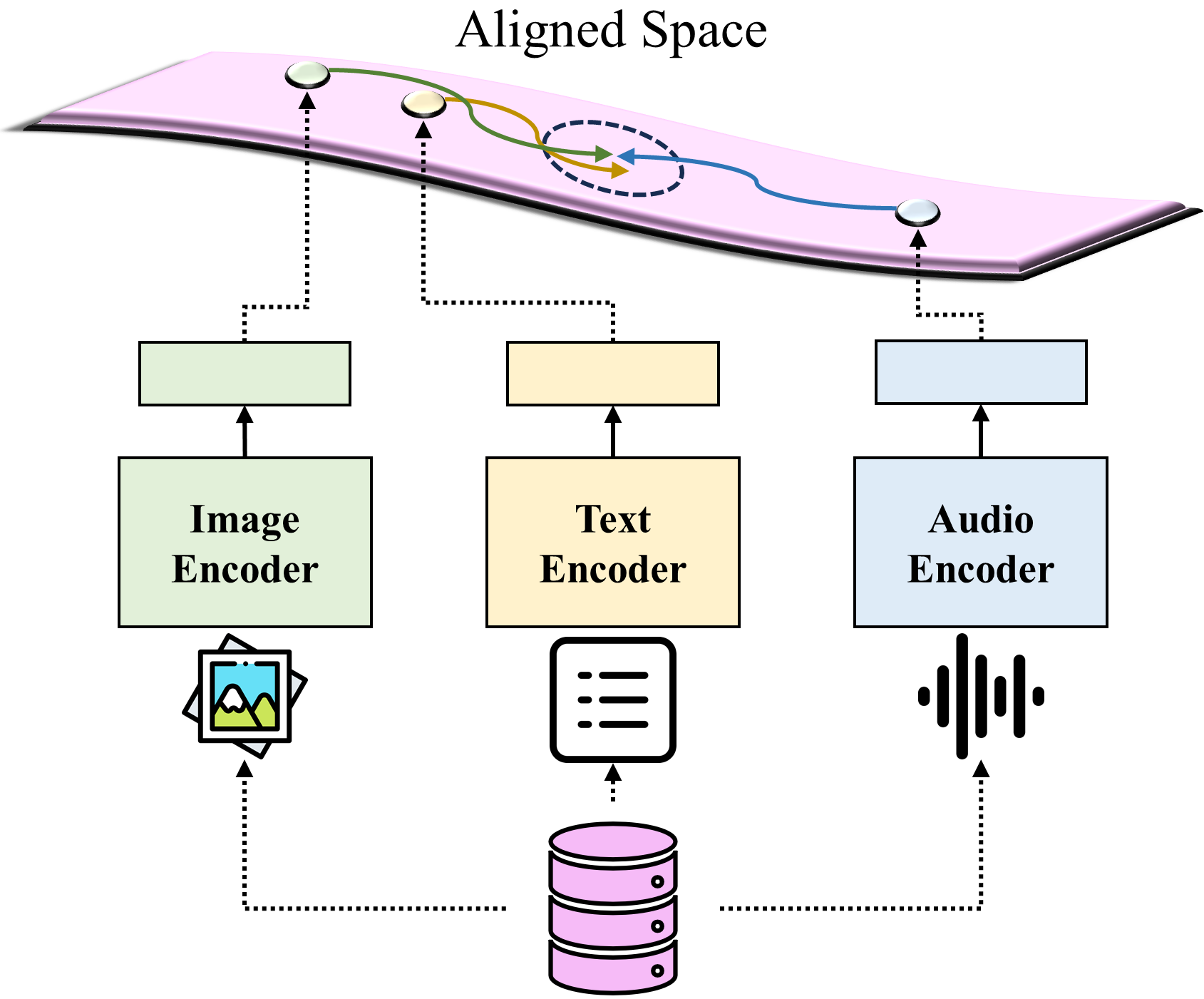} 
\caption{\textbf{Mapping to aligned space for cross-modality.} The aligned space acts as a convergence point for correlating different modal inputs and leveraging them for further multi-modal processing.}
\label{fig:Intro}
\end{figure}

Motivated by the these limitations and the need for studies to process and integrate diverse information like humans, we introduce Synergy-CLIP, a novel framework for tri-modal representation learning. The objectives of Synergy-CLIP are illustrated in figure~\ref{fig:Intro}. It aligns and captures the latent information embedded within intricately intertwined tri-modal data. 
In addition, we propose the VGG-sound$^{\text{+}}$ dataset to pre-train Synergy-CLIP. The VGG-sound$^{\text{+}}$ dataset is a tri-modal dataset built upon VGG-sound \cite{chen2020vggsound}, containing matched visual-audio-text data. It includes image-text-audio modalities by augmenting the visual-audio pairs with appropriate textual descriptions, resulting in a balanced dataset where all three modalities are represented at equal scale.
%To pre-train Synergy-CLIP, we construct a VGG-sound$^{\text{+}}$ dataset by enhancing the VGG-sound dataset \cite{chen2020vggsound}. The VGG-sound$^{\text{+}}$ dataset includes audio-visual data augmented with appropriate text descriptions, thus encompassing tri-modal image-text-audio data. 
Further, we utilize specific modules of the trained Synergy-CLIP to perform various validations. We perform the Missing Modality Reconstruction (MMR) task using learned high-dimensional representations to evaluate the effectiveness of Synergy-CLIP in capturing synergies between modalities and reconstructing missing modalities in various scenarios. Additionally, we validate the representational power of Synergy-CLIP using modality-specific downstream tasks, including zero-shot classification. The experimental results confirm that the proposed methodology outperforms the baseline in certain tasks.

This study makes the following contributions to multi-modal learning:
\begin{itemize}
    \item \textbf{Introduction and explanation of the tri-modal dataset:} This study introduces a comprehensive tri-modal dataset, VGG-sound$^{\text{+}}$, which is expected to serve as a valuable resource for future research on tri-modal learning scenarios.
    \item \textbf{Proposal of Synergy-CLIP:} We propose Synergy-CLIP, a new model that extends the traditional CLIP framework to account for and process tri-modal data, creating a foundation for research to leverage diverse information in a human-like manner.
    \item \textbf{Introduction of a new validation methodology:} To validate the effectiveness of Synergy-CLIP, we introduce the MMR task. By evaluating the model's ability to reconstruct missing modal information, a more realistic validation metric is introduced for performance evaluation in real-life applications, engendering new perspectives on multi-modal learning.
\end{itemize}
This research not only paves the way for more sophisticated integration of multi modalities within AI systems, but also highlights the potential for leveraging different data types to understand and solve complex problems. Additionally, by leveraging MMR tasks, it will be possible to develop a wide range of applications in real-world domains, such as healthcare and natural sciences.

The remainder of this paper is organized as follows. Section~\ref{sec:Related_works} reviews related work on multi-modal representation learning and missing modality reconstruction. Section~\ref{sec3} introduces the Synergy-CLIP framework and the VGG-sound+ dataset. Section~\ref{sec4} presents experimental results, highlighting Synergy-CLIP's performance. Section~\ref{sec:conclusion} concludes with contributions and future directions, and Section~\ref{sec:limitation} discusses limitations and ethics.

%%%%%%%%%%%%%%%%%%%%%%%%%%%%%%%%%%%%%%%%%%%%%%%%%%%%%%%%%%%%%%%%%%%%%%%%

\section{Related work}\label{sec:Related_works}

\subsection{multi-modal Representation Learning}
Several studies have been conducted to understand and integrate multi-modal data within deep learning frameworks \cite{baltruvsaitis2018multimodal,yang2017deep}. These studies have made significant contributions in the field of multi-modal representation learning, which seeks to build integrated representations from various data sources, such as images, text, and audio data, or images from RGB-Depth-Segment \cite{wang2018depth,bachmann2022multimae} modalities. Existing studies have primarily utilized transformer-based architectures \cite{lu2019vilbert,al2023vision} to capture complex relationships based on inter-modality attention, exhibiting exceptional predictive power. Recently, CLIP has been introduced as a groundbreaking approach \cite{radford2021learning}, which uses separate encoders for images and text based on contrastive learning to extract representations for each modality and align them. CLIP exhibits effective representation extraction performance by leveraging aligned representations, enabling zero-shot tasks to be performed based on integrated expressions. The ability of CLIP to utilize large datasets for training not only bridges the semantic gap between visual and textual data, but also unveils new directions for subsequent research in this field \cite{elizalde2023clap,zhang2022pointclip,guzhov2022audioclip,tschannen2023clippo,wu2024vadclip}.

Building on the foundation established by CLIP, recent research has demonstrated the potential of broadening the horizons of multi-modal representation learning comprehensively by exploring combinations of modalities beyond traditional image-text pairs, such as audio-text \cite{elizalde2023clap,guzhov2022audioclip} and 3D image-text \cite{zhang2022pointclip} pairs. In this context, we aim to extend existing methodologies by simultaneously connecting not just two but three modalities. Previous research on expanding from bimodal to tri-modal analysis has focused on "adapting" \cite{ruan2023accommodating, wu2022wav2clip} representations of newly added modalities using the robustly trained vanilla-CLIP. However, this does not constitute an equitable expansion of the modality scale. This study emphasizes the necessity and importance of a dataset that represents the three most widely used data modalities, image, text, and audio, equally. Thus, our goal is to process and learn data from these three modalities on an equal footing to achieve high-dimensional representation learning that integrates and understands the modalities more comprehensively. Further, we investigate the impact of high-level integrated representations corresponding to various subtasks and modality combinations. The proposed Synergy-CLIP framework and our investigation enhances the understanding and integration of multi-modal data and suggests new directions for future research in the field of AI.

\subsection{Missing Modality Reconstruction}
Recently, considerable research has been conducted to address the problem of missing modalities in multi-modal data \cite{ma2021smil,woo2023towards}. Previous studies have either advocated for learning shared features \cite{wang2023multi} across modalities or used Transformer Encoders to understand the relationships between missing modalities based on attention matrices \cite{ma2022multimodal}, reporting improvements in robustness. In addition, methods such as Semi-Supervised Learning and Knowledge Distillation \cite{maheshwari2024missing} have been used to learn representations of missing modalities from teacher models. Most previous studies have aimed to bridge the information gap induced by missing modalities during training or inference stages \cite{ma2022multimodal,maheshwari2024missing}. Although effective, these approaches result in unconventional and complex architectures \cite{wang2023multi}. In contrast, we propose an MMR task that aims to emulate human cognitive abilities by reconstructing missing modalities using available ones. This task focuses on reconstruction, which differs significantly from existing methodologies in terms of its objectives for handling missing modalities. We utilize high-dimensional integrated representations constructed using the Synergy-CLIP architecture to reconstruct the features of the missing modality, and ultimately, the missing modality itself. To achieve this goal, we introduce a transformer-based encoder-decoder framework. The structure of the proposed model is illustrated in figure~\ref{fig:MMR_archi}.

We consider three reconstruction scenarios based on the constructed Triplet Dataset (image-text-audio).
\begin{enumerate}
    \item {\textbf{Missing Image Reconstruction.} When the vision modality is missing, the proposed model uses only text and audio modalities to reconstruct the missing visual content. This approach demonstrates the model's capability to depict objects described by text-audio information visually.}
    \item {\textbf{Missing Text Reconstruction.} When the text modality is missing, the proposed model uses the image and audio modalities to reconstruct a description or label that represents the provided content and context accurately. This showcases the model's ability to articulate what it ``sees'' and ``hears'' clearly.}
    \item {\textbf{Missing Audio Reconstruction.} When the audio modality is missing, the proposed model uses the image and text modalities to reconstruct the missing auditory content. This demonstrates the model's ability to express auditory information clearly based on the provided visual content and descriptions.}
\end{enumerate}   
By proposing an approach for the reconstruction of multi-modal data, we introduce a new perspective on existing research directions and provide a foundation for the ultimate goals of AI technology, that is, emulating human cognitive abilities. Moreover, this design enhances the robustness of models concerning incomplete data and expands their applicability to scenarios in which data are inherently partial or subject to various limitations.

% Intro Figure
% Intro Figure

%%%%%%%%%%%%%%%%%%%%%%%%%%%%%%%%%%%%%%%%%%%%%%%%%%%%%%%%%%%%%%%%%%%%%%%%

% Synergy-Clip Figure
\begin{figure*}[!t]
\centering
\includegraphics[width=0.985\textwidth]{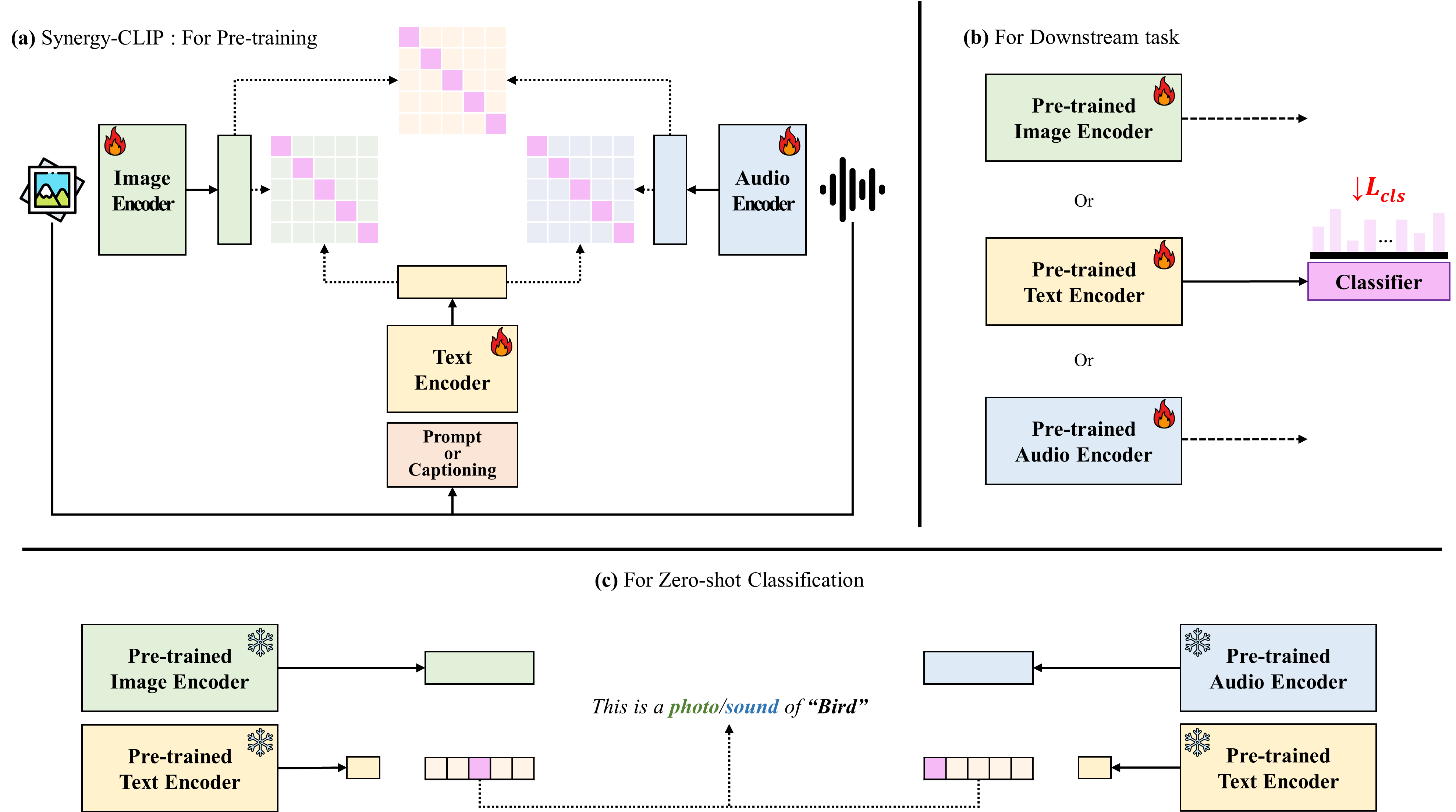} 
\caption{\textbf{Illustration of Synergy-CLIP framework}. (a) Pre-training of Synergy-CLIP involves inputs from three modalities: image, text, and audio. During pre-training, representations extracted by each modality-specific encoder are aligned, utilizing contrastive loss to facilitate this alignment. The pre-trained encoders of Synergy-CLIP are subsequently fine-tuned for modality-specific downstream tasks (b) as well as zero-shot classification (c), enabling the framework to adaptively leverage aligned representations across various AI tasks.}
\label{fig:synergy_clip}
\end{figure*}
% Synergy-Clip Figure

\section{Synergy-CLIP}\label{sec3}

\subsection{VGG-sound$^{\text{+}}$ Benchmark Construction}
We construct the triple-modality dataset, VGG-sound$^{\text{+}}$, comprising image-text-audio data to pre-train Synergy-CLIP. VGG-sound \cite{chen2020vggsound} is a large-scale dataset designed for audio-visual learning, encompassing a diverse range of environmental and action sounds across various categories. It is widely utilized in the multimedia (audio-visual) domain and comprises over 310 distinct sound classes, including musical instruments, animal sounds, and mechanical noises. Each sample consists of an audio-visual pair extracted from YouTube videos, making it a valuable resource for multi-modal learning and retrieval tasks. Based on VGG-sound, VGG-sound$^{\text{+}}$ consists of \textit{200,000} audio-visual data entries categorized as video data, including metadata labeling the category of each video clip. We define the image-text-audio triplet modalities of VGG-sound$^{\text{+}}$ as the dataset $D_i = (I_i, T_i, A_i)$, where $I_i$ represents an image snapshot of the video, $T_i$ denotes a textual description of the video, and $A_i$ signifies the audio clip. $D_i$ is utilized during pre-training and on MMR tasks.

\textbf{Adding a text description.} We extract random scenes from approximately 10 seconds-long video clips for $I_i$ and capture the audio of the video clips for $A_i$. However, the VGG-sound dataset contains only metadata specifying the category of each video clip, without accompanying textual captions describing them in detail. Thus, we generate suitable captions $T_i$ for the acquired $I_i$ and $A_i$. Two approaches to captioning are considered: the semi-handcrafted method and the captioning model method. Inspired by previous image-text pre-training studies \cite{radford2021learning}, the semi-handcrafted approach constructs text description sets by randomly combining categories recorded in metadata with appropriate prompt templates (e.g., \textit{'a photo and sound of [category]'}). We devise 71 prompts for this training method. In contrast, the captioning model method uses extracted images and audio as inputs to generate text descriptions that include words recorded in the metadata. We use BLIP-2 \cite{li2023blip} as the captioning model. The equation of the text description method, $\text{TD}(\cdot)$, for each $T_i$ is as follows:
\begin{equation}
    \label{eq:text_descript}
    \begin{aligned}
        T_i = 
        \begin{cases} 
        \text{TD}_{P}(\text{template}, \text{category}) & \text{if method is } P \\
        \text{TD}_{C}(I_i, A_i, \text{metadata}) & \text{if method is } C
        \end{cases}
    \end{aligned}
\end{equation}
where $P$ denotes the semi-handcrafted approach and $C$ represents the captioning model method. We anticipate that text descriptions generated using the captioning model provide the model with richer expressiveness.

\subsection{Tri-modal representation learning}
We conduct pre-training using the constructed image-text-audio triplets to align the representations of each modality. In particular, we implement a model structure \cite{dosovitskiy2020image,gong2021ast,liu2019roberta} that performs well corresponding to each modality and use it as a representation encoder for each modality. We define the representations to be aligned for each modality as follows:
\begin{equation}
    \label{eq:modal_features}
    \begin{aligned}
    h_i^{img} &= \mathsf{F}_{img}(I_i) \\
    h_i^{txt} &= Avg\_pool(\mathsf{F}_{txt}(T_i)) \\
    h_i^{aud} &= \mathsf{F}_{aud}(A_i)
    \end{aligned}
\end{equation}
where $\mathsf{F}_{img}(\cdot), \mathsf{F}_{txt}(\cdot)$, and $\mathsf{F}_{aud}(\cdot)$ denote the encoders for the different modalities, and $h_i^{img}, h_i^{txt}$, and $h_i^{aud}$ denote the respective modality representations extracted by these encoders.

\textbf{Aligning Representations of Triplet Modalities.} For each object $i$, $I_i, T_i$, and $A_i$ denote the inputs into their respective encoders used to extract the representations, $h_i^{img}, h_i^{txt}$, and $h_i^{aud}$, respectively. These are aligned using the following equation:

\begin{equation}
    \label{eq:pretrain_objective_function}
    \begin{aligned}
    \mathcal{L}_{total} &= \alpha\mathcal{L}_{clip}(h^{img}, h^{txt}) \\
    &+ \beta\mathcal{L}_{clip}(h^{txt}, h^{aud}) \\
    &+ \gamma\mathcal{L}_{clip}(h^{aud}, h^{img})
    \end{aligned}
\end{equation}

where $\alpha, \beta$, and $\gamma$ denote hyper-parameters used to align the modality representations, with default values set to 1 for maintain balance among modality combinations. If $\alpha$ and $\beta$ are 1 and $\gamma$ is 0.5, the audio-image modality would be learned less effectively. Synergy-CLIP aims to capture interactions among modalities under equal conditions. Testing to prevent image-text over-fitting with $\alpha$ at 0.5 and $\beta$, $\gamma$ at 1 showed inferior results, supporting the hypothesis for equal settings. A detailed analysis of the experimental results is provided in the Ablation Study section. Further, $\mathcal{L}_{clip}(\cdot)$ denotes the contrastive loss for paired modality representations. The detailed equation is as follows:

\begin{equation}
    \label{eq:clip_loss}
%    \resizebox{.6\hsize}{!}{%
    \begin{aligned}
    \mathcal{L}_{clip}(h^{M_1}, h^{M_2}) = &-\frac{1}{N} \sum_{i=1}^{N} \left[ \log \frac{\exp\left(\frac{\textit{sim}(h^{M_1}_i, h^{M_2}_i)}{\tau}\right)}{\sum_{j}^{N} \exp\left(\frac{\textit{sim}(h^{M_1}_i, h^{M_2}_j)}{\tau}\right)} \right. \\
    &+ \left. \log \frac{\exp\left(\frac{\textit{sim}(h^{M_1}_i, h^{M_2}_i)}{\tau}\right)}{\sum_{j}^{N} \exp\left(\frac{\textit{sim}(h^{M_1}_j, h^{M_2}_i)}{\tau}\right)} \right]
    \end{aligned}
%}
\end{equation}

where $M_1$ and $M_2$ are paired modalities, and $h$ represents the modality vectors. The function $\textit{sim}(\cdot)$ computes the similarity between two representations, and $\tau$ denotes a temperature parameter used to scale the similarity scores. $\mathcal{L}_{clip}$ maximizes the cosine similarity between correct modality pairs and minimizes it between incorrect pairs, thereby learning semantic relationships between the modalities.

% MMR Figure
\begin{figure}[!t]
\centering
\includegraphics[width=0.975\columnwidth]{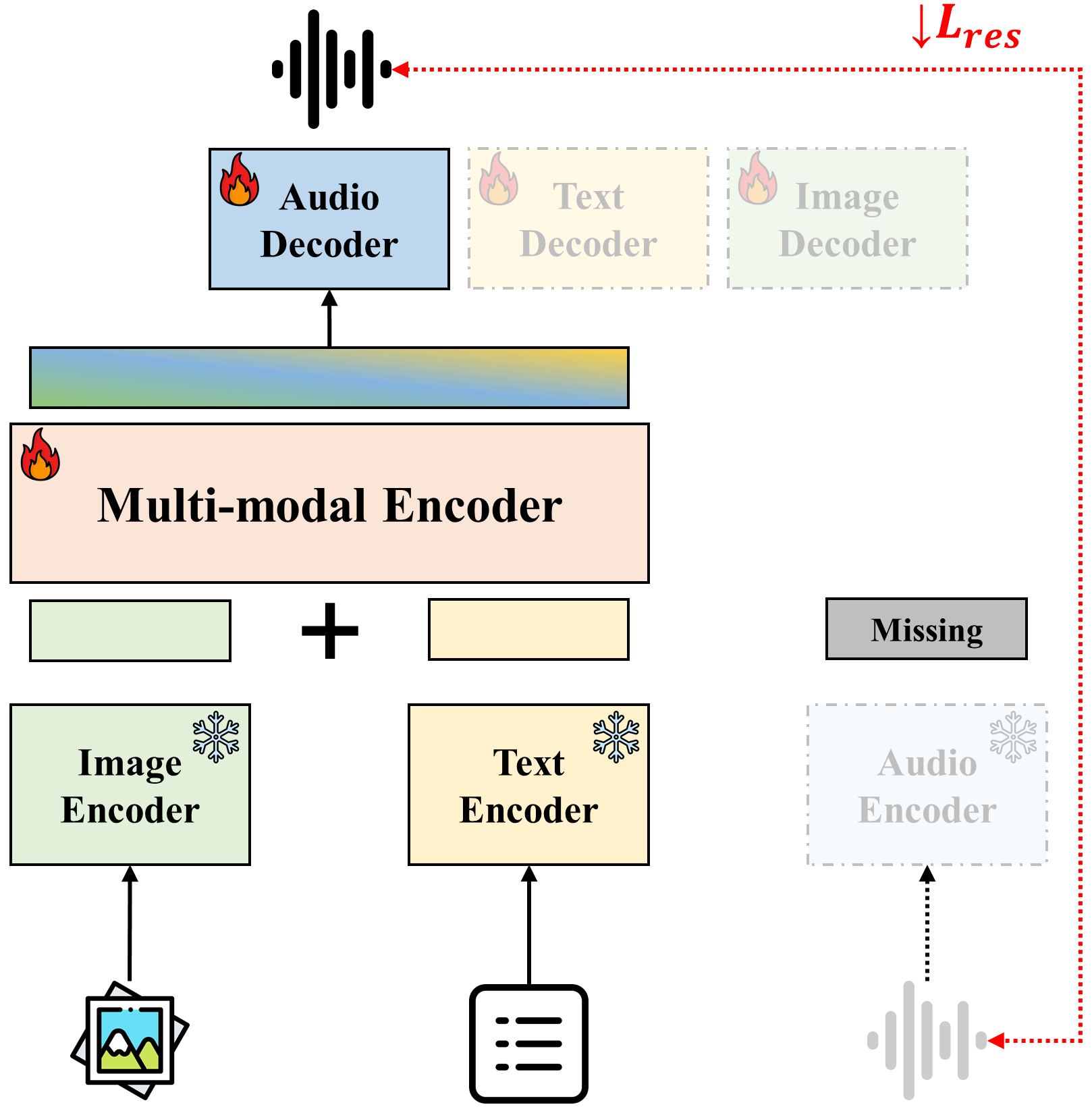} 
\caption{\textbf{Framework for the Missing Modality Reconstruction (MMR) task}, this figure illustrates the use of a multi-modal encoder and a missing modality decoder to reconstruct the audio modality using representations extracted by Synergy-CLIP.}
% \caption{Example framework for performing the \textbf{Missing Modality Reconstruction(MMR)} task (in case of audio modality missing). Design a combination of Multi-modal Encoder and Missing modal Decoder utilizing pre-trained Synergy-CLIP as a feature extractor. At training time, we build a model as a scenario for the missing modality to represent the features of the missing modality through the remaining modalities and reconstruct them through the modality decoder.}
\label{fig:MMR_archi}
\end{figure}
% MMR Figure

\subsection{Missing Modality Reconstruction}
The primary aim of AI development is to emulate human cognitive capabilities to develop versatile and powerful systems \cite{brynjolfsson2022turing}. One of the remarkable abilities of humans is to perceive and reconstruct missing modalities using available sensory data. For instance, hearing a bird chirping might inspire the composition of a beautiful poetic verse or conjure a vivid visualization of the bird itself. To emulate such connections, we define and perform MMR, enabling AI systems to infer and reconstruct absent modalities. Although MMR is conceptually straightforward, its realization faces a complex challenge. Given a set of modalities, the goal is to reconstruct a missing modality using the information provided by the available modalities. By learning to infer and reconstruct absent forms, the proposed model enhances its resilience to data incompleteness, thereby improving its robustness and adaptability in real-world applications where data incompleteness or corruption are common.

The proposed model structure for performing MMR is illustrated in figure~\ref{fig:MMR_archi}. As illustrated in the figure, we design a multi-modal encoder and a missing modal decoder to execute MMR. For each modality, that is, image, text, and audio, representations are initially obtained using the pre-trained encoders. Subsequently, one modality input is masked, and the model is used to encode information about the missing modality using the remaining ones. For instance, if the audio modality is missing (see figure~\ref{fig:MMR_archi}), the model relies on the image and text representations to encode information about the missing audio modality. The representation of the missing modality is then transmitted to the decoder, which reconstructs the modality. The model is trained by minimizing the loss $\mathcal{L}_{res}$ between the original and reconstructed forms.

\textbf{Objective Function For MMR.} To optimize MMR, we define the following objective function:
\begin{equation}
    \label{eq:recon_loss}
    \begin{aligned}
    \mathcal{L}_{\text{res}}(M, \hat{M}) =
    \begin{cases} 
    \mathcal{L}_{res}^{img}(M, \hat{M},\delta), & \text{if } M \text{ is } I \\
    \mathcal{L}_{res}^{txt}(M, \hat{M},\eta), & \text{if } M \text{ is } T \\
    \mathcal{L}_{res}^{aud}(M, \hat{M},\theta), & \text{if } M \text{ is } A
    \end{cases}
    \end{aligned}
\end{equation}
where $\mathcal{L}_{\text{res}}(M, \hat{M})$ is calculated dynamically for the image ($I$), text ($T$), and audio ($A$) modalities. Additionally, $M$ represents the original modality and $\hat{M}$ represents the modality reconstructed by the model.

If the image ($I$) modality is missing, the objective function becomes $\mathcal{L}_{res}^{img}(I, \hat{I},\delta)$. $\mathcal{L}_{res}(\cdot)$ is composed of $\text{SSIM}$ and $\ell_2$, where $\text{SSIM}$ captures the essential structural information present in visual data, and $\ell_2$ preserves pixel-level details. The complete expression for image reconstruction is as follows:
\begin{equation}
    \label{eq:recon_loss_img}
    \begin{aligned}
    \mathcal{L}_{res}^{img}(I, \hat I,\delta) &= \delta \cdot \text{SSIM}(I, \hat{I}) \\
    &+ (1 - \delta) \cdot \mathcal{L}_{\ell_2}(I, \hat{I})
    \end{aligned}
\end{equation}
where the default value of $\delta$ is set to 0.75. This is because a greater weight needs to be assigned to $\text{SSIM}$ during image reconstruction as it captures colorful details more effectively than $\ell_2$-loss.

If the text ($T$) modality is missing, the \textit{ cross entropy loss } loss function is employed, which is effective in the context of categorical distributions in textual data. The loss function is expressed as follows:
\begin{equation}
    \label{eq:recon_loss_text}
    \begin{aligned}
    \mathcal{L}_{res}^{txt}(M, \hat M, \eta) = \eta \cdot \text{CE-Loss}(T, \hat{T})
    \end{aligned}
\end{equation}
where the default value of $\eta$ is 1.0.

If the audio ($A$) modality is missing, the objective function is taken to be $\mathcal{L}_{res}^{aud}(A, \hat{A},\theta)$. $\mathcal{L}_{res}^{aud}(\cdot)$ includes $\text{SSIM}$ and $\ell_2$. As in the case of images, $\text{SSIM}$ captures the structural aspects of audio waveforms while $\ell_2$-loss ensures the preservation of waveform pixel values. The complete expression for audio reconstruction is as follows:
\begin{equation}
    \label{eq:recon_loss_aud}
    \begin{aligned}
    \mathcal{L}_{res}^{aud}(A, \hat A,\theta) & = \theta \cdot \text{SSIM}(A, \hat{A}) \\
    & + (1 - \theta) \cdot \mathcal{L}_{\ell_2}(A, \hat{A})
    \end{aligned}
\end{equation}
where the default value of $\theta$ is 0.25. In contrast to image reconstruction, audio reconstruction involves only a single channel. Therefore, $\theta$ is set to prioritize $\ell_2$loss, which excels at approximating the pixel values themselves.

During MMR training, the gradients are managed such that the weights of the modality-specific encoders, which are learned during the pre-training process, are frozen. Only the multi-modal encoder and the missing modal decoder are updated.

%%%%%%%%%%%%%%%%%%%%%%%%%%%%%%%%%%%%%%%%%%%%%%%%%%%%%%%%%%%%%%%%%%%%%%%%

\section{Experiments}\label{sec4}
In this section, we evaluate the performance of the proposed Synergy-CLIP on various downstream tasks, including zero-shot classification, and assess the MMR solution performance of the multi-modal encoder and missing modal decoder frameworks based on Synergy-CLIP.

\subsection{Implementation Details}
\textbf{During pre-training}, we implement the $\text{Synergy-CLIP}_{base}$ and $\text{Synergy-CLIP}_{large}$ models based on each backbone network (see figure~\ref{fig:synergy_clip} top-left). Additionally, text descriptions are generated using semi-handcrafted prompts ($P$) and BLIP-2 \cite{li2023blip} captions ($C$) during pre-training. We use four A6000 GPUs for training. For detailed pre-training settings, please refer to table~\ref{tab:pt_set}. Reproducibility was ensured by fixing the Seed at $17$.

\begin{table}[!h]
\centering
\caption{Default Pre-training Setting}
\label{tab:pt_set}
\resizebox{0.665\columnwidth}{!}{%
\begin{tabular}{@{}ll@{}}
\toprule
\textbf{Hyper-parameters}                         & \textbf{Value}       \\ \midrule
Optimizer                                         & AdamW                \\
learning rate                                     & 5e-6                 \\
Weight decay                                      & 1e-5                 \\
Adam $\beta$                                        & (0.9, 0.999)         \\
Batch size                                        & 256                  \\
Training epochs                                   & 10                   \\
projection dim                                    & 1024                 \\
$\alpha$ in equation~\ref{eq:pretrain_objective_function} & 1.0                  \\
$\beta$ in equation~\ref{eq:pretrain_objective_function}  & 1.0                  \\
$\gamma$ in equation~\ref{eq:pretrain_objective_function} & 1.0                  \\ \midrule
Image resolution                                  & 224 x 224            \\
Image augmentation                                & RandomResizedCrop    \\
                                                  & RandomHorizontalFlip \\
                                                  & RandomVerticalFlip   \\ \midrule
Text max length                                   & 32                   \\ \midrule
Audio sample rate                                 & 16000                \\
Audio Augmentation                                & AddWhiteNoise        \\
                                                  & Shifting             \\
                                                  & Stretching           \\
                                                  & Flipping             \\ \bottomrule
\end{tabular}%
}
\end{table}

\textbf{For Missing Modality Reconstruction (MMR) task}, a multi-modal encoder and missing modal decoder-based framework is devised based on pre-trained Synergy-CLIP to reconstruct the missing modalities. The encoder is designed to fuse representations of the remaining modalities. The decoders for image and audio uses a CNN-based structure, and the text decoder is based on a transformer-based structure. We also construct and train the models for various missing scenarios based on model size and text description (see figure~\ref{fig:MMR_archi}). We use a pair of A6000 GPUs. For detailed MMR-training settings, please refer to table~\ref{tab:mmt_set}. Reproducibility was maintained with a fixed seed of $42$.

\begin{table}[!t]
\centering
\caption{Default MMR task Setting}
\label{tab:mmt_set}
\resizebox{0.5\columnwidth}{!}{%
\begin{tabular}{@{}ll@{}}
\toprule
\textbf{Hyper-parameters}        & \textbf{Value} \\ \midrule
Optimizer                        & AdamW          \\
learning rate                    & 1e-4           \\
Weight decay                     & 1e-5           \\
Adam $\beta$                       & (0.9, 0.999)   \\
Batch size                       & 128            \\
Training epochs                  & 200            \\
projection dim                   & 1024           \\
$\delta$ in equation~\ref{eq:recon_loss}  & 0.75           \\
$\eta$ in equation~\ref{eq:recon_loss}    & 1.0            \\
$\theta$ in equation~\ref{eq:recon_loss} & 0.25           \\ \midrule
Image resolution                 & 224 x 224      \\ \midrule
Text max length                  & 32             \\ \midrule
Audio sample rate                & 16000          \\ \bottomrule
\end{tabular}%
}
\end{table}

\begin{table}[!b]
\centering
\caption{Default Fine-tuning \& Zero-shot Setting}
\label{tab:ft_set}
\resizebox{0.675\columnwidth}{!}{%
\begin{tabular}{@{}ll@{}}
\toprule
\textbf{Hyper-parameters} & \textbf{Value}       \\ \midrule
Optimizer                 & AdamW                \\
learning rate             & 1e-5                 \\
Weight decay              & 1e-5                 \\
Adam $\beta$              & (0.9, 0.999)         \\
Batch size                & 256                  \\
Training epochs           & 10                   \\ \midrule
Image resolution          & 224 x 224            \\
Image augmentation        & RandomResizedCrop    \\
                          & RandomHorizontalFlip \\
                          & RandomVerticalFlip   \\ \midrule
Text max length           & 32                   \\ \midrule
Audio sample rate         & 16000                \\
Audio Augmentation        & AddWhiteNoise        \\
                          & Shifting             \\
                          & Stretching           \\
                          & Flipping             \\ \bottomrule
\end{tabular}%
}
\end{table}

%%%%%%%%%%%%%%%%%%%%%%%%%%% MMR Table %%%%%%%%%%%%%%%%%%%%%%%%%%%
\begin{table*}[!t]
	\centering
    \caption{\textbf{Performance of MMR.} This table presents the reconstruction performance in various scenarios with missing modalities. Training reports on randomly sampled subsets of triplet datasets are listed, which contain similar meta-information at a rate of 10\% of the total data.}
    \label{tab:mmr_table}
	\resizebox{\textwidth}{!}{%
		% \scalebox{1.135}{%
        \begin{tabular}{@{}cc|ccc|ccc|ccc@{}}
        \toprule
        \multicolumn{2}{c|}{\textbf{Modality}}   & \multicolumn{3}{c|}{\textbf{Image}}                       & \multicolumn{3}{c|}{\textbf{Text}}                            & \multicolumn{3}{c}{\textbf{Audio}}                      \\
        \textbf{Captioning}      & \textbf{Size} & \textbf{LPIPS(↓)} & \textbf{PSNR (↑)} & \textbf{SSIM (↑)} & \textbf{METEOR(↑)} & \textbf{FENSE(↑)} & \textbf{Accuracy(↑)} & \textbf{MCD(↓)} & \textbf{PSNR (↑)} & \textbf{SSIM (↑)} \\ \midrule
        \multirow{2}{*}{Prompt}  & Base          & 0.17              & 21.30             & 0.89              & 55.86              & 66.08             & 67.32                & 5.19            & 14.17             & 0.67              \\
                                 & Large         & 0.15              & 22.19             & 0.92              & 57.53              & 68.17             & 73.54                & 4.41            & 12.07             & 0.88              \\
        \multirow{2}{*}{Caption} & Base          & 0.17              & 21.32             & 0.89              & 57.32              & 69.36             & 75.46                & \textbf{3.25}   & \textbf{14.36}    & 0.87              \\
                                 & Large         & \textbf{0.14}     & \textbf{22.27}    & \textbf{0.92}     & \textbf{58.07}     & \textbf{69.98}    & \textbf{75.62}       & 3.38            & 14.34             & \textbf{0.92}     \\ \bottomrule
        \end{tabular}
		}
	\end{table*}
%%%%%%%%%%%%%%%%%%%%%%%%%%% MMR Table %%%%%%%%%%%%%%%%%%%%%%%%%%%

\textbf{In downstream tasks}, we also perform experiments to validate the expressive power of pre-trained Synergy-CLIP. Common downstream tasks, primarily in the fields of vision, text, and audio, are considered for validation. Each modality-specific encoder is fine-tuned along with a classifier to perform tasks ranging from conventional classification to zero-shot classification (see figure~\ref{fig:synergy_clip} top right \& bottom). For each task, we fixed Seeds at three values ($17$, $42$, $77$) to provide an average of performance and its standard deviation. For detailed pre-training settings, please refer to table~\ref{tab:ft_set}.

\subsection{Backbone Networks}
We configure the following backbone networks for each modality:

\noindent\textbf{For the vision model}, the ViT-base\cite{dosovitskiy2020image} and ViT-large architectures are chosen owing to their proven efficiency in capturing complex visual patterns using a self-attention mechanism, which considers task-related image areas flexibly.

\noindent\textbf{For the text model}, we use the RoBERTa-base\cite{liu2019roberta} and RoBERTa-large architectures. Owing to more focused pre-training using dynamic masking, these models are effective in capturing complex text and integrated features over various modalities, and they outperform BERT\cite{devlin2018bert}.

\noindent\textbf{For the audio model}, we utilize AST\cite{gong2021ast}, whose Base and Large models differ not in the number of parameters but in the overall patch size. AST is also a transformer-based architecture, which is capable of modeling and capturing complex acoustic patterns effectively.

\subsection{Datasets}
For each modality, we evaluate the representations of pre-trained Synergy-CLIP on the following downstream tasks:

\noindent\textbf{Image Datasets,} Oxford-IIIT Pets\cite{parkhi12a}, a dataset of images featuring 37 common breeds of pets, with approximately 200 images per breed, is primarily used to evaluate fine-grained visual classification algorithms. Flowers-102\cite{Nilsback08}, composed of images of 102 types of commonly seen flowers, with each category containing between 40 and 258 images, is used as a challenging dataset for fine-grained image recognition tasks. Subsets of a labeled portion from a collection of 80 million tiny images, CIFAR\cite{krizhevsky2009learning}, CIFAR-10, containing 6,000 images for each of 10 classes, and CIFAR-100, containing 600 images for each of 100 classes, are used to evaluate image recognition algorithms.

\noindent\textbf{Text Datasets,} GLUE Dataset\cite{wang2018glue}, the General Language Understanding Evaluation (GLUE) benchmark, which aggregates various natural language understanding tasks, includes tasks for evaluating and analyzing model performance across various linguistic phenomena, such as sentiment analysis, question answering, and text implication.

\noindent\textbf{Audio Datasets,} ESC-50\cite{piczak2015dataset}, a dataset composed of short, labeled environmental sound clips spread over 50 classes, serves as a standardized dataset for benchmarking models for environmental sound classification. It is also used to compare various methodologies in audio recognition tasks. We follow the dataset's recommendations for training/validation using 5-fold cross-validation. UrbanSound8k\cite{Salamon:UrbanSound:ACMMM:14}, a dataset containing 8,732 labeled short sound clips recorded in urban environments, is classified into ten classes. It is extensively used to evaluate the performance of urban sound classification algorithms, emphasizing the importance of audio signal processing in machine learning. We follow the dataset's recommendations for training/validation using 10-fold cross-validation.

%%%%%%%%%%%%%%%%%%%%%%%%%%% Image FT Table %%%%%%%%%%%%%%%%%%%%%%%%%%%
\begin{table*}[!t]
	\centering
    \caption{This table presents the performance of the image classification downstream tasks. Synergy-CLIP improves upon the performance of baselines within the same CLIP series on some tasks. The average values are reported along with the corresponding standard deviations over three different seed values.}
    \label{tab:image_ft}
	\resizebox{0.80\textwidth}{!}{%
		% \scalebox{1.115}{%
        \begin{tabular}{@{}cccccc@{}}
        \toprule
        \multicolumn{1}{l}{}                                     & \textbf{Oxford-IIIT Pets} & \textbf{Flowers-102}  & \textbf{CIFAR-10}     & \textbf{CIFAR-100}    & \textbf{avg}          \\ \midrule
        ResNet 152 \cite{radford2021learning}   & 93.00                     & 89.60                 & 93.50                 & 78.00                 & 88.53                 \\
        BiT-M \cite{kolesnikov2020big}          & 92.40                     & 99.30                 & 97.60                 & 88.20                 & 94.38                 \\
        ViT-B \cite{dosovitskiy2020image}       & 90.40                     & 98.70                 & 96.70                 & 86.30                 & 93.03                 \\
        ViT-L \cite{dosovitskiy2020image}       & 92.90                     & 99.30                 & 97.90                 & 89.00                 & 94.78                 \\
        CLIP : ViT-B \cite{radford2021learning} & 90.00                     & 96.90                 & 95.10                 & 80.50                 & 90.63                 \\
        CLIP : ViT-L \cite{radford2021learning} & 95.10                     & 99.20                 & 98.00                 & 87.50                 & 94.95                 \\ \midrule
        Synergy-CLIP                                             & \multicolumn{5}{c}{Ours}                                                                                                  \\
        Base\_\{P\} : vision enc                                 & 91.99 ± 0.05              & 97.68 ± 0.02          & 97.31 ± 0.01          & 83.28 ± 0.04          & 92.57 ± 0.03          \\
        Base\_\{C\} : vision enc                                 & 91.76 ± 0.01              & 97.22 ± 0.04          & 97.10 ± 0.01          & 87.25 ± 0.03          & 93.33 ± 0.02          \\
        Large\_\{P\} : vision enc                                & 94.40 ± 0.02              & \textbf{99.54 ± 0.01} & 98.22 ± 0.01          & \textbf{90.82 ± 0.02} & \textbf{95.74 ± 0.01} \\
        Large\_\{C\} : vision enc                                & \textbf{94.73 ± 0.01}     & 98.80 ± 0.08          & \textbf{98.39 ± 0.02} & 90.74 ± 0.02          & 95.67 ± 0.03          \\ \bottomrule
        \end{tabular}
		}
	\end{table*}
	%%%%%%%%%%%%%%%%%%%%%%%%%%% Image FT Table %%%%%%%%%%%%%%%%%%%%%%%%%%%

\begin{table*}[!t]
\centering
\caption{This table presents the results of the GLUE Benchmark. In certain tasks, Synergy-CLIP exhibits superior performance compared to pre-trained models of similar size. The best performance out of the three seed-trained outcomes is reported here.}
\label{tab:text_ft}
\resizebox{0.85\textwidth}{!}{%
\begin{tabular}{@{}cccccccccc@{}}
\toprule
\multicolumn{1}{l}{} &
  \textbf{MNLI-M} &
  \textbf{MNLI-MM} &
  \textbf{QNLI} &
  \textbf{QQP} &
  \textbf{SST-2} &
  \textbf{CoLA} &
  \textbf{MRPC} &
  \textbf{RTE} &
  \textbf{avg} \\ \midrule
BiLSTM+Attn, ELMo                         & 72.40 & 72.40 & 75.20 & 83.60          & 91.50 & 44.10 & 82.10 & 52.70 & 71.75 \\
Bert-base\cite{devlin2018bert}            & 84.00 & 84.20 & 91.00 & 87.60          & 92.60 & 60.30 & 90.20 & 69.50 & 82.43 \\
RoBERTa\cite{liu2019roberta}              & 90.20 & 90.20 & 94.70 & 92.20          & 96.40 & 68.00 & 90.90 & 86.60 & 88.65 \\
GPT-2 XL\cite{radford2019language}        &   -   & 71.50 & 76.70 & 70.10          & 91.30 & 10.50 & 69.40 & 56.80 & 63.76 \\
GPT-3\cite{brown2020language}             &   -   & 90.20 & 92.70 & 81.10          & 97.10 & 60.30 & 86.50 & 83.70 & 84.51 \\
CLIP : txt enc.\cite{tschannen2023clippo} & 71.80 & 72.50 & 73.00 & 82.70          & 86.20 & 6.60  & 81.40 & 53.80 & 66.00 \\ \midrule
\textbf{Synergy-CLIP}                     & \multicolumn{9}{c}{\textbf{Ours}}                                              \\
Base$_{P\text{ : txt enc.}}$                & 87.31 & 87.04 & 91.60 & 90.10          & 93.12 & 69.12 & 90.09 & 61.73 & 83.76 \\
Base$_{C\text{ : txt enc.}}$                & 87.53 & 87.31 & 91.93 & 90.11          & 93.35 & 69.13 & 89.30 & 65.70 & 84.30 \\
Large$_{P\text{ : txt enc.}}$               & 90.23 & 90.40 & 94.45 & \textbf{91.74} & 96.10 & 83.89 & 89.56 & 74.01 & 88.80 \\
Large$_{C\text{ : txt enc.}}$ &
  \textbf{90.29} &
  \textbf{90.49} &
  \textbf{94.75} &
  91.71 &
  \textbf{96.22} &
  \textbf{85.23} &
  \textbf{90.80} &
  \textbf{79.42} &
  \textbf{89.86} \\ \bottomrule
\end{tabular}%
}
\end{table*}
\begin{table}[!b]
	\centering
    \caption{This table presents the results achieved in the sound classification downstream task. Synergy-CLIP outperforms the baseline in some tasks. Results are reported as the average value with the standard deviation over three seed values. The best performance compared to the baselines is marked with an asterisk\textbf{(*)}.}
    \label{tab:audio_ft}
	\resizebox{\columnwidth}{!}{%
		% \scalebox{0.80}{%
			\begin{tabular}{@{}ccc@{}}
				\toprule
				\multicolumn{1}{l}{} & \textbf{ESC-50}                & \textbf{UrbanSound8k}          \\ \midrule
				Human                & 81.50                          & -                              \\
				ESResNeXt\cite{guzhov2021esresne}            & 95.20                          & 89.14                          \\
				AST\cite{gong2021ast}                  & 95.60                          & -                              \\
				AudioCLIP\cite{guzhov2022audioclip}            & 97.15                          & 90.07                          \\
				CLAP\cite{elizalde2023clap}                 & 96.70                          & 87.96                          \\ \midrule
				\textbf{Synergy-CLIP}         & \multicolumn{2}{c}{\textbf{Ours}}                               \\
				Base$_{P\text{ : aud enc.}}$    & 95.05 ± 0.02 \textbf{(*97.75)}          & \textbf{89.31 ± 0.03} (*94.44) \\
				Base$_{C\text{ : aud enc.}}$   & 94.55 ± 0.01 (*96.75)          & 89.18 ± 0.02 \textbf{(*94.55)}          \\
				Large$_{P\text{ : aud enc.}}$  & 94.70 ± 0.02 (*96.75)          & 87.83 ± 0.04 (*94.02)          \\
				Large$_{C\text{ : aud enc.}}$  & \textbf{95.10 ± 0.02} (*96.75) & 88.50 ± 0.03 (*91.45)          \\ \bottomrule
			\end{tabular}%
			}
	\end{table}
%%%%%%%%%%%%%%%%%%%%%%%%%%% Audio FT Table %%%%%%%%%%%%%%%%%%%%%%%%%%%

\subsection{Missing Modality Reconstruction}
In this section, we validate the performance of the proposed methodology in the MMR task both quantitatively and qualitatively. 

\textbf{Quantitative evaluations.} Table~\ref{tab:mmr_table} presents the quantitative evaluations of the encoder-decoder structure for the Multi-Modal Reconstruction (MMR) task, assessing the reconstruction performance across different missing-modality scenarios. For each modality, we employ appropriate evaluation metrics to measure the quality of the reconstructed outputs. The results highlight that models pre-trained with captioned data consistently outperform those trained with handcrafted prompts. Furthermore, increasing the model size (\textit{base} → \textit{large}) leads to a noticeable improvement in reconstruction performance across all modalities.
For scenarios where the \textbf{image modality is missing}, we evaluate the reconstructed images using Peak Signal-to-Noise Ratio (PSNR) \cite{hore2010pnsr}, Structural Similarity Index Measure (SSIM) \cite{wang2004image}, and Learned Perceptual Image Patch Similarity (LPIPS). LPIPS \cite{zhang2018unreasonable} measures perceptual similarity, where lower values indicate higher reconstruction quality. As shown in Table~\ref{tab:mmr_table}, larger models demonstrate superior reconstruction performance, achieving higher PSNR and SSIM scores while reducing LPIPS. The caption-based large model achieves the best image reconstruction results (PSNR: 22.27, SSIM: 0.92, LPIPS: 0.14), surpassing both prompt-based and base-sized models.
In cases where the \textbf{text modality is missing}, we assess the reconstructed text using METEOR \cite{banerjee2005meteor} and FENSE \cite{zhang2019bertscore} scores, as well as accuracy, which measures the alignment between the reconstructed and ground-truth tokens. As observed in Table~\ref{tab:mmr_table}, models trained with captioned data exhibit stronger text reconstruction performance compared to those trained with handcrafted prompts. The caption-based large model achieves the highest scores (METEOR: 58.07, FENSE: 69.98, Accuracy: 75.62).
For \textbf{missing audio modality} scenarios, we evaluate reconstruction quality using Mel Cepstral Distortion (MCD) \cite{kominek2008synthesizer} (where lower values indicate better quality) along with PSNR and SSIM. The results reveal that models pre-trained with captions outperform those trained with prompts, with the caption-based large model achieving the lowest MCD (3.38) and the highest PSNR (14.34) and SSIM (0.92). This demonstrates that caption-based training facilitates better alignment between audio and other modalities, leading to improved reconstruction quality.

\begin{figure*}[!t]
\centering
\includegraphics[width=\textwidth]{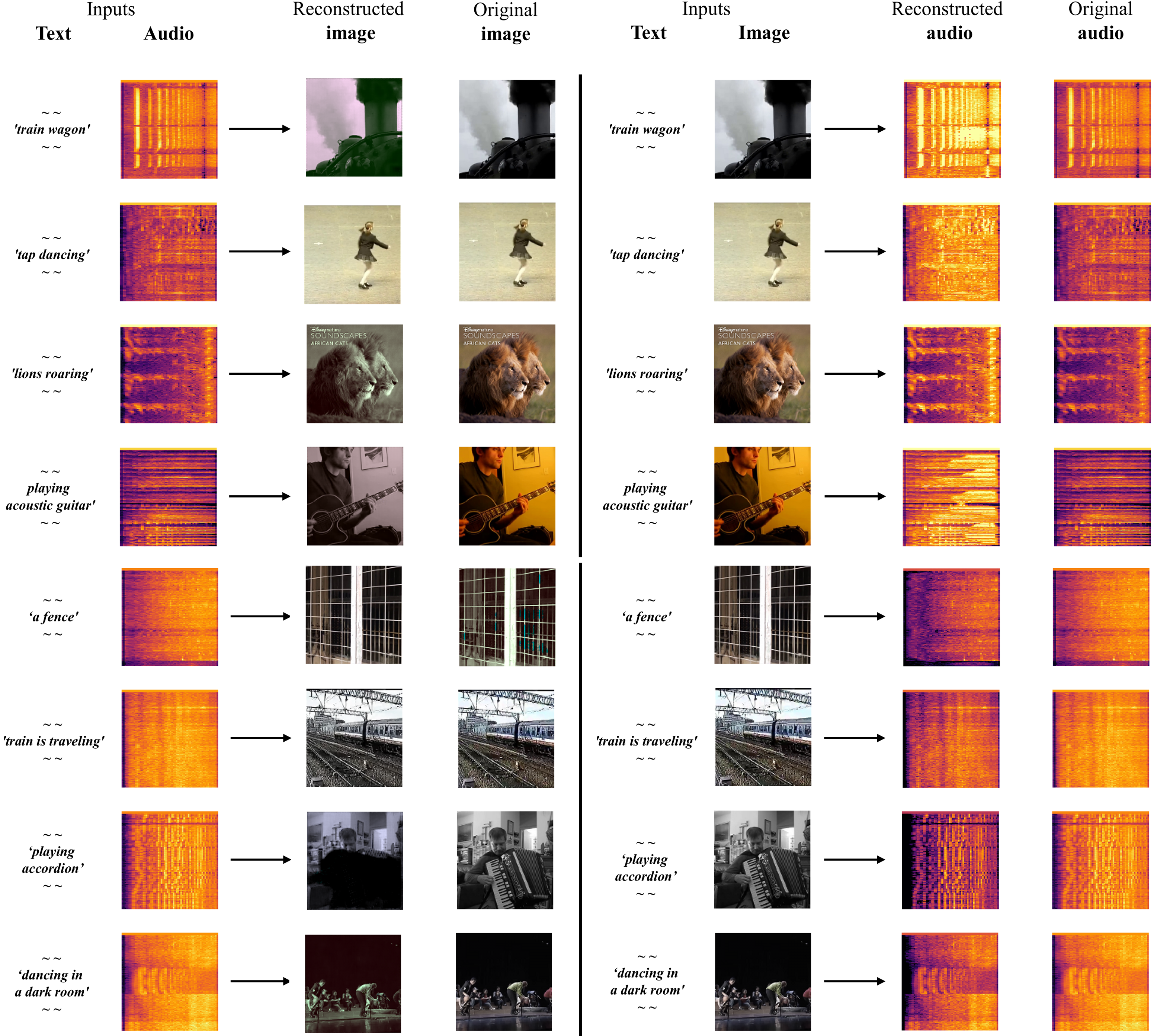} 
\caption{\textbf{A qualitative example of the MMR} task. In this evaluation, the text reconstruction scenario has been excluded (as it is a token classification). The figure illustrates the reconstruction results of the large caption model, which exhibits the best performance as per quantitative evaluations.} 
\label{fig:MMR_deploy}
\end{figure*}

\textbf{Qualitative evaluation.} We also present a qualitative evaluation in figure~\ref{fig:MMR_deploy}. The figure~\ref{fig:MMR_deploy} illustrates the inference results in MMR scenarios obtained using the best-performing Large-Caption model. Both image and audio results are observed to be highly similar to the actual outputs. The color differences likely result from the impact of hyper-parameters in equation~\ref{eq:recon_loss_img} and~\ref{eq:recon_loss_aud} and the reconstruction performance of the decoders. Adjusting the boundaries of $\delta$ and $\theta$ in equation~\ref{eq:recon_loss_img} and~\ref{eq:recon_loss_aud}, which balance SSIM and L2 loss, is necessary. SSIM loss emphasizes brightness, contrast, and luminance, causing certain parts of reconstructed modality to be highlighted differently. Also, We observed that the model produced outputs for both images and audio that closely resembled real examples across various categories. However, in certain samples, reconstruction failed, particularly with more complex objects such as musical instruments. This suggests that cross-modal representations in audio or text are weaker in these cases, highlighting a key challenge for MMR to address in future work. Nonetheless, the model demonstrates strong qualitative performance in reconstruction overall.

%%%%%%%%%%%%%%%%%%%%%%%%%%% Image-Text ZS Table %%%%%%%%%%%%%%%%%%%%%%%%%%%
\begin{table*}[!t]
\centering
\caption{This table presents the zero-shot classification performance of Synergy-CLIP for image-text  modality, with respect to the number of modalities aligned during pre-training. The data indicate that aligning the representations of three modalities yields better outcomes compared to aligning just two. Further, pre-training with captions instead of simple prompts yields enhanced representations.}
\label{tab:zero_shot_img}
\resizebox{0.85\textwidth}{!}{%
% \scalebox{1.075}{%
\begin{tabular}{@{}ccccccc@{}}
\toprule
\multirow{2}{*}{\textbf{Captioning Module}} &
  \multirow{2}{*}{\textbf{Size}} &
  \multirow{2}{*}{\textbf{Modality aligned}} &
  \multicolumn{2}{c}{\textbf{CIFAR-10}} &
  \multicolumn{2}{c}{\textbf{CIFAR-100}} \\
                         &                        &           & \textbf{top - 1}      & \textbf{top - 5}      & \textbf{top - 1}      & \textbf{top - 5}      \\ \midrule
\multirow{4}{*}{Prompt}  & \multirow{2}{*}{Base}  & I + T     & 76.04 ± 0.04          & 99.74 ± 0.04          & 19.01 ± 0.04          & 41.15 ± 0.08          \\
                         &                        & I + T + A & 76.56 ± 0.06          & 98.43 ± 0.01          & 22.01 ± 0.03          & 45.83 ± 0.03          \\
                         & \multirow{2}{*}{Large} & I + T     & 84.04 ± 0.01          & 98.96 ± 0.03          & 24.23 ± 0.02          & 49.22 ± 0.03          \\
                         &                        & I + T + A & \textbf{84.38 ± 0.01} & \textbf{99.22 ± 0.01} & \textbf{28.26 ± 0.02} & \textbf{54.04 ± 0.02} \\ \midrule
\multirow{4}{*}{Caption} & \multirow{2}{*}{Base}  & I + T     & 69.53 ± 0.02          & 98.44 ± 0.01          & 16.92 ± 0.03          & 38.80 ± 0.02          \\
                         &                        & I + T + A & 77.60 ± 0.02          & 97.66 ± 0.01          & 24.48 ± 0.03          & 50.13 ± 0.04          \\
                         & \multirow{2}{*}{Large} & I + T     & 70.57 ± 0.03          & 98.18 ± 0.01          & 21.35 ± 0.02          & 50.26 ± 0.07          \\
                         &                        & I + T + A & \textbf{86.20 ± 0.05} & \textbf{99.22 ± 0.01} & \textbf{31.51 ± 0.03} & \textbf{61.07 ± 0.04} \\ \bottomrule
\end{tabular}%
}
\end{table*}
%%%%%%%%%%%%%%%%%%%%%%%%%%% Image-Text ZS Table %%%%%%%%%%%%%%%%%%%%%%%%%%%

%%%%%%%%%%%%%%%%%%%%%%%%%%% Audio-Text ZS Table %%%%%%%%%%%%%%%%%%%%%%%%%%%
% Please add the following required packages to your document preamble:
% \usepackage{booktabs}
% \usepackage{multirow}
% \usepackage{graphicx}
\begin{table*}[!t]
\centering
\caption{This table presents the zero-shot classification performance of Synergy-CLIP for audio-text  modality. As with table~\ref{tab:zero_shot_img}, we see that aligning the three modalities performs better, and the Caption has a better representation.}
\label{tab:zero_shot_aud}
\resizebox{0.85\textwidth}{!}{%
% \scalebox{1.075}{%
\begin{tabular}{@{}ccccccc@{}}
\toprule
\multirow{2}{*}{\textbf{Captioning Module}} &
  \multirow{2}{*}{\textbf{Size}} &
  \multirow{2}{*}{\textbf{Modality aligned}} &
  \multicolumn{2}{c}{\textbf{ESC-50}} &
  \multicolumn{2}{c}{\textbf{UrbanSound8k}} \\
                         &                        &           & \textbf{top - 1}      & \textbf{top - 5}      & \textbf{top - 1}      & \textbf{top - 5}      \\ \midrule
\multirow{4}{*}{Prompt}  & \multirow{2}{*}{Base}  & A + T     & 56.25 ± 0.07          & 85.45 ± 0.02          & 48.56 ± 0.03          & 84.38 ± 0.01          \\
                         &                        & A + T + I & \textbf{66.88 ± 0.04} & 88.59 ± 0.02          & 52.81 ± 0.06          & \textbf{89.84 ± 0.02} \\
                         & \multirow{2}{*}{Large} & A + T     & 56.25 ± 0.02          & 86.72 ± 0.05          & 51.56 ± 0.05          & 84.22 ± 0.02          \\
                         &                        & A + T + I & 65.47 ± 0.04          & \textbf{88.75 ± 0.03} & \textbf{53.98 ± 0.05} & 85.23 ± 0.06          \\ \midrule
\multirow{4}{*}{Caption} & \multirow{2}{*}{Base}  & A + T     & 44.79 ± 0.05          & 76.82 ± 0.02          & 39.32 ± 0.03          & 88.80 ± 0.01          \\
                         &                        & A + T + I & 59.84 ± 0.04          & 85.63 ± 0.03          & 46.48 ± 0.08          & 85.08 ± 0.03          \\
                         & \multirow{2}{*}{Large} & A + T     & 55.83 ± 0.02          & 78.13 ± 0.01          & 51.30 ± 0.08          & \textbf{92.96 ± 0.05} \\
                         &                        & A + T + I & \textbf{66.25 ± 0.02} & \textbf{87.66 ± 0.03} & \textbf{58.59 ± 0.05} & 90.00 ± 0.03          \\ \bottomrule
\end{tabular}%
}
\end{table*}
%%%%%%%%%%%%%%%%%%%%%%%%%%% Audio-Text ZS Table %%%%%%%%%%%%%%%%%%%%%%%%%%%

\subsection{Experiments based on Downstream Tasks}

\subsubsection{Single-modal Results}
We perform and validate downstream tasks for each modality. The results for image, text, and audio data are presented in table~\ref{tab:image_ft},~\ref{tab:text_ft} and~\ref{tab:audio_ft}, respectively. 

\noindent \textbf{In image tasks.} we found that Synergy-CLIP outperformed the baseline and a model trained in a similar context(CLIP\cite{radford2021learning} series). In particular, the Large models achieved the highest performance on all datasets. This means that Synergy-CLIP has learned a more sophisticated feature extraction process.

\noindent \textbf{In text tasks.} Moreover, the proposed model outperforms the baseline. In addition, it outperforms RoBERTa \cite{liu2019roberta} on average, despite statistically insignificant differences on some tasks. This confirms that Synergy-CLIP has the ability to generalize across a variety of natural language understanding tasks.

\noindent \textbf{In audio tasks.} Finally, the proposed model outperforms similarly trained models (AudioCLIP\cite{guzhov2022audioclip} and CLAP\cite{elizalde2023clap}). These results underscore the model's ability to capture nuanced audio features, further supporting its effectiveness in extracting modality-specific representations.

Overall, the downstream task evaluations across the three modalities validate Synergy-CLIP's capability to enhance performance across diverse tasks, achieving state-of-the-art results or competitive performance. The model’s ability to outperform specialized models in each modality—image, text, and audio—highlights its potential for multi-modal applications, where robust feature extraction and representation are crucial. The consistency of high performance across different datasets and tasks further substantiates Synergy-CLIP as an effective multi-modal learning framework.
% The results of the downstream tasks on the three modalities confirm that the proposed Synergy-CLIP framework extracts higher-level representations and features than other baseline models.

\subsubsection{Zero-shot Classification Results}
Finally, we conduct and validate zero-shot classification for the previously performed downstream tasks. Zero-shot classification is highly influenced by the scale and quality of the pre-training data; therefore, to ensure fair evaluation, we report the results using models pre-trained in the same way as the previously introduced data. The performance of Synergy-CLIP in zero-shot classification is presented in table~\ref{tab:zero_shot_img} and~\ref{tab:zero_shot_aud}. The \textit{``Modality aligned''} column in tables lists the results with respect to combinations of modalities with aligned representations during pre-training. As indicated in the table~\ref{tab:zero_shot_img},~\ref{tab:zero_shot_aud}, Synergy-CLIP, which aligns all three modalities, generally exhibits superior performance. Additionally, pre-training with captions, rather than simple prompts, yields enhanced expressiveness. These results confirm that the number of modalities aligned impacts the ability to capture latent relationships among them significantly and that using well-articulated captions over simple prompts enhances the expressive power of CLIP-series models.

\begin{table*}[!t]
\centering
\caption{Impact of different $\alpha, \beta, \gamma$ values on R@10 performance. The baseline configuration ($\alpha = \beta = \gamma = 1.0$) achieves the highest overall R@10 score (90.06\%), indicating that balanced weight settings yield the best modality alignment. Reducing any of the weights individually (A1, A2, A3) leads to a drop in performance, especially in the text-audio and audio-image retrieval tasks. When multiple weights are reduced simultaneously (B1, B2, B3), the decline becomes more significant, confirming the necessity of equal weighting for optimal retrieval performance.}
\label{tab:ablation}
\resizebox{0.675\textwidth}{!}{%
\begin{tabular}{cccccccc}
\toprule
& $\alpha$ & $\beta$ & $\gamma$ & R@10$^{\text{img-txt}}$ & R@10$^{\text{txt-aud}}$ & R@10$^{\text{aud-img}}$ & R@10$^{\text{avg}}$ \\
\midrule
Baseline & 1.0 & 1.0 & 1.0 & \textbf{96.25} & \textbf{89.65} & \textbf{84.29} & \textbf{90.06} \\
\midrule
A1 & 0.5 & 1.0 & 1.0 & 94.17 & 87.23 & 83.17 & 88.19 \\
A2 & 1.0 & 0.5 & 1.0 & 95.72 & 83.96 & 83.29 & 87.66 \\
A3 & 1.0 & 1.0 & 0.5 & 94.79 & 87.44 & 81.12 & 87.78 \\
\midrule
B1 & 1.0 & 0.5 & 0.5 & 95.85 & 82.92 & 82.28 & 87.02 \\
B2 & 0.5 & 1.0 & 0.5 & 94.27 & 88.62 & 81.98 & 88.29 \\
B3 & 0.5 & 0.5 & 1.0 & 94.38 & 82.44 & 83.94 & 86.92 \\
\bottomrule
\end{tabular}
}
\end{table*}

\subsection{Ablation Study on Modality Alignment}
A controlled study was conducted to analyze the effects of the pre-training objective function defined in Eq.\ref{eq:pretrain_objective_function}. The experiments primarily focused on two aspects: the impact of reducing each hyperparameter individually and the effect of jointly decreasing two hyperparameters. After pre-training on VGG-sound$^{\text{+}}$, we evaluated the R@10 performance on the test subset of VGG-sound$^{\text{+}}$. The detailed results are provided in Table\ref{tab:ablation}.

In the experiments where one of $\alpha$, $\beta$, or $\gamma$ was individually adjusted (A1, A2, A3), a noticeable decline in R@10 was observed, highlighting the importance of balanced hyperparameter settings. Specifically, as the alignment strength between certain modalities weakened due to altered training parameters, the overall performance deteriorated accordingly.

Moreover, the experiments in which two of $\alpha$, $\beta$, or $\gamma$ were simultaneously reduced (B1, B2, B3) exhibited even more pronounced performance degradation. These results clearly demonstrate that weakening the alignment between specific modalities negatively impacts the alignment and performance of other modalities as well.

The findings indicate that maintaining equal weighting across all modalities yields the most optimal performance. When any of $\alpha$, $\beta$, or $\gamma$ was reduced, the R@10 performance dropped significantly, with the most substantial decline occurring in text-audio and audio-image retrieval. These results further substantiate the critical role of balanced modality alignment in ensuring robust multi-modal learning.

%%%%%%%%%%%%%%%%%%%%%%%%%%%%%%%%%%%%%%%%%%%%%%%%%%%%%%%%%%%%%%%%%%%%%%%%

% MMR deploy Figure
% MMR deploy Figure

\section{Conclusion and Future Work} \label{sec:conclusion}
In this study, we propose Synergy-CLIP, an extension of the CLIP framework designed to enhance the expressiveness of models across visual, textual, and audio modalities and emulate human cognitive abilities. By extending the modalities and framework, we overcome the limitations of existing studies on paired modalities and achieve significant advancements. We also introduce the Missing Modality Reconstruction (MMR) task and perform comprehensive evaluations on various downstream tasks, including zero-shot classification. The results further demonstrate the robustness and flexibility of Synergy-CLIP.

This study emphasizes the importance of a holistic approach to multi-modal learning, in which coordinating and integrating multiple modalities can significantly improve model performance in various tasks. In particular, zero-shot classification results reveal that models pre-trained with more nuanced and comprehensive data (e.g., captions rather than simple prompts) exhibit superior performance, underscoring the importance of rich and descriptive pre-training in enhancing model functionality. Additionally, the MMR task, involving the reconstruction of missing sensory information based on available data, serves as a unique metric for assessing the resilience of models in real-world application environments. It not only tests the limits of current AI capabilities but also establishes a foundation for future research on understanding and interpreting incomplete or partial data.

The learning approach of Synergy-CLIP can be extended to incorporate a broader range of modalities, further enhancing the model's sensory integration capabilities. Additionally, by advancing the MMR task, the model can capture more nuanced interactions across modalities, thereby supporting various real-world applications through the reconstruction of missing information. Future research will focus on securing a richer diversity of modality data to reinforce inter-modality interactions robustly and on refining the MMR task to elevate its applicability in practical domains such as healthcare and security. Furthermore, the introduction of additional modalities will enable the exploration of a wider range of missing modality reconstruction scenarios, allowing Synergy-CLIP to handle more complex and diverse real-world cases. By addressing these challenges, future iterations of the model aim to further enhance its adaptability and effectiveness across multiple domains.

Overall, Synergy-CLIP sets a new benchmark for accommodating diverse modalities, potentially having a significant impact on multi-modal representational learning and serving as a pivotal foundation for understanding the complex interplay between various modalities.

\section{Limitations and Ethics}\label{sec:limitation}
\textbf{Limitations.} While Synergy-CLIP demonstrates robust performance leveraging multi-modal datasets, it also presents limitations under specific conditions and assumptions. In terms of scalability, adding new modalities requires extensive resources for constructing and maintaining large-scale datasets that equally represent each modality. Furthermore, the additional loss term in equation~\ref{eq:pretrain_objective_function} increases computational costs, potentially constraining practical scalability. Specifically, when introducing a new modality, the pre-training objective function incurs an 
$\comb{n}{2}$ computational overhead due to pairwise modality alignment, significantly increasing complexity as the number of modalities grows. Additionally, in the Missing Modality Reconstruction (MMR) task, Synergy-CLIP may struggle to reconstruct complex data, potentially impacting performance in applications that require precise restoration. Another limitation arises from BLIP-generated captions, which, while effective for multi-modal alignment, can be susceptible to inherent biases within the BLIP model. These biases may propagate into the training process, affecting the quality and diversity of textual representations. Addressing these biases is crucial to ensuring fair and unbiased learning across different modalities.

\textbf{Ethics.} The ethical considerations surrounding Synergy-CLIP must be carefully addressed in both research and real-world applications. First, there is the issue of data bias. As Synergy-CLIP trains on large multi-modal datasets, it may inherit latent biases present within these datasets, which could reinforce or amplify societal biases in sensitive application areas. Continuous monitoring and dataset review during model training are essential to mitigate this risk. Additionally, the use of BLIP-generated captions introduces another layer of potential bias. Since BLIP itself is trained on large-scale vision-language datasets, it may exhibit biases in how it describes images, leading to skewed or stereotypical textual representations that can propagate through Synergy-CLIP’s training pipeline. Careful curation and bias auditing of generated captions are necessary to prevent unintended reinforcement of these biases.

Second, there are privacy concerns. The MMR task, which infers missing visual or audio data based on existing information, could pose privacy risks in contexts where personal or public system data privacy is critical. In particular, the ability to reconstruct missing modalities raises concerns about potentially inferring sensitive or unintended information from incomplete data. Consequently, Synergy-CLIP applications, particularly those involving MMR, should operate under strict ethical oversight, ensuring that the model does not inadvertently infer sensitive or personally identifiable information.

\bibliographystyle{unsrt}
\bibliography{reference}

\EOD

\end{document}